\documentclass[12pt]{article}

\usepackage[utf8]{inputenc}
\usepackage[T1]{fontenc}
\usepackage[english]{babel}
\usepackage[top=1in, bottom=1in, left=1in, right=1in]{geometry}
\usepackage{setspace}
\usepackage{fancyhdr}
\newcommand{\runningheads}[2]{
    \setlength{\headheight}{15pt}
    \addtolength{\topmargin}{-2pt}
    \pagestyle{fancy}  
    \fancyhead{}  
    \fancyhead[L]{\small #1}
    \fancyhead[R]{\small #2}
    \renewcommand{\headrulewidth}{0.4pt}
}

\usepackage{float}
\usepackage{graphicx}
\usepackage{subcaption}
\usepackage{booktabs}
\usepackage{makecell}
\usepackage{multirow}
\usepackage{threeparttable}  

\usepackage[linesnumbered,ruled,vlined]{algorithm2e}
\SetCommentSty{textnormal}
\usepackage{enumitem}
\usepackage{pdfpages}

\usepackage{amsmath}
\usepackage{amssymb}
\usepackage{amsthm}

\theoremstyle{definition}

\usepackage{authblk}

\usepackage{natbib}
\usepackage{xcolor}
\definecolor{darkblue}{rgb}{0, 0, 0.75}
\definecolor{darkgreen}{rgb}{0, 0.75, 0}
\usepackage[colorlinks=true, linkcolor=black, citecolor=darkgreen, urlcolor=darkblue]{hyperref}

\newenvironment{foldable}{}{}
\newcommand*{\ie}{i.e.}
\newcommand*{\eg}{e.g.}

\newcommand{\meanstd}[2]{#1$_{\pm{#2}}$}
\newcommand{\meanstdbf}[2]{\textbf{#1}$_{\pm{#2}}$}
\newcommand{\boldtight}[1]{{\fontseries{b}\selectfont #1}}
\newcommand*{\bltrg}{$^{\blacktriangle}$}

\title{Diverse Image Priors \\ for Black-box Data-free Knowledge Distillation}
\author[1]{Tri-Nhan Vo}
\author[1]{Dang Nguyen}
\author[2]{Trung Le}
\author[1]{Kien Do}
\author[1]{Sunil Gupta}
\affil[1]{Applied Artificial Intelligence Initiative, Deakin University, Geelong, Australia}
\affil[1]{\texttt{\{s223032975, d.nguyen, kien.do, s.gupta\}@deakin.edu.au}}
\affil[2]{Department of Data Science \& AI, Monash University, Melbourne, Australia}
\affil[2]{\texttt{\{trunglm\}@monash.edu.au}}
\runningheads{T-N. Vo et al.}{Diverse Image Priors for Black-box Data-free KD}
\date{\small{March 2026}}

\begin{document}

\maketitle

\begin{abstract} \label{sec:00-abstract}
  Knowledge distillation (KD) represents a vital mechanism to transfer expertise from complex teacher networks to efficient student models.
However, in decentralized or secure AI ecosystems, privacy regulations and proprietary interests often restrict access to the teacher's interface and original datasets.
These constraints define a challenging black-box data-free KD scenario where only top-1 predictions and no training data are available.
While recent approaches utilize synthetic data, they still face limitations in data diversity and distillation signals.
We propose \textbf{Diverse Image Priors Knowledge Distillation} (\textbf{DIP-KD}), a framework that addresses these challenges through a three-phase collaborative pipeline: (1) \textit{Synthesis} of image priors to capture diverse visual patterns and semantics; (2) \textit{Contrast} to enhance the collective distinction between synthetic samples via contrastive learning; and (3) \textit{Distillation} via a novel primer student that enables soft-probability KD.
Our evaluation across 12 benchmarks shows that DIP-KD achieves state-of-the-art performance, with ablations confirming data diversity as critical for knowledge acquisition in restricted AI environments.


\end{abstract}

\section{Introduction} \label{sec:01-introduction}
\begin{foldable} 
  The evolution of Artificial Intelligence has resulted in high-capacity models that serve as centralized repositories of specialized expertise.
  However, the substantial size of these models often limits their mobility and deployment in decentralized environments, such as mobile or edge devices.
  To bridge this gap, research on model compression seeks to transfer these expert insights into lightweight, portable agents.
  \textit{Knowledge distillation} (KD)~\cite{hinton2015distilling} is the primary framework for this pedagogical exchange, where an expert network (teacher) guides the learning of a novice network (student).
\end{foldable}

\begin{foldable} 
  In practical AI ecosystems, this information flow is frequently obstructed by systemic barriers that prevent transparent knowledge exchange.
  Stringent privacy regulations, such as HIPAA or GDPR~\cite{us1996hipaa, eu2016gdpr}, prohibit access to original training data, particularly in sensitive domains, such as personal, health, and proprietary data.
  Furthermore, intellectual property or security concerns~\cite{us2016dsta} often render the teacher as \textit{black-box}, withholding internal signals like activations, gradients, or class probabilities, and returning only the top-1 predictions.
  These constraints define the \textit{black-box data-free KD} (BBDFKD) problem: training a student to emulate an expert's intelligence through a narrow interface without any historical data.
\end{foldable}

\begin{foldable} 
  Recent methods in BBDFKD~\cite{wang2021zero,zhang2022ideal,yuan2024data} leverage synthetic data, but they still encounter two primary bottlenecks.
  First, they suffer from a lack of domain coverage: synthetic samples fail to replicate the hierarchical structures and semantic variation of natural environments, resulting in a distribution mismatch with the teacher's expertise.
  Second, the distillation signals are suboptimal.
  By relying exclusively on hard top-1 indices, students are denied the \textit{dark} knowledge of inter-class relationships---a cornerstone of KD essential for emulating expert logic.
  Within a collective intelligence framework, this resembles an agent attempting to reconstruct a diverse collective memory from discrete feedback, deprived of the historical data that shaped the expert's knowledge.
\end{foldable}

\begin{foldable} 
  \textbf{Our method.} We propose \textbf{Diverse Image Priors Knowledge Distillation} (\textbf{DIP-KD}) to tackle the BBDFKD problem, \ie, training a student from a black-box teacher with only top-1 signal and without access to any real data.
  To improve KD, we incorporate \textit{diversity}, \ie, how broadly the samples distribute, into a collaborative three-phase pipeline:{
    (1)~\textit{Synthesis}: crafts synthetic \textit{image priors} with diverse patterns and semantics of natural objects;
    (2)~\textit{Contrast}: employs a \textit{primer student} to act as a white-box mediator and ensure synthetic samples are informationally distinct via contrastive learning;
    and (3)~\textit{Distillation}: distills the final student optimally with hard labels from teacher and soft probabilistic labels from the primer student.
  }
\end{foldable}

\begin{foldable} 
  \textbf{Contribution.} In summary, we highlight our contributions as follows:
  \begin{enumerate}
    \item We introduce \textit{image priors}, a novel class of synthetic images exhibiting diversity across hierarchical structures, nonlinearity, and semantics.
    \item We propose the novel use of a \textit{primer student}, enabling contrastive optimization and extraction of soft probabilistic signals originally restricted in black-box KD.
    \item Extensive analysis of our framework demonstrates that it effectively expands the semantic coverage of synthetic domains, leading to our state-of-the-art performance across 12 benchmarks.
  \end{enumerate}
\end{foldable}

\section{Related Works} \label{sec:02-related}
\subsection{Foundations of Knowledge Transfer} \label{ssec:02-related-kd}
\begin{foldable} 
  The pioneering idea in knowledge transfer between neural networks, intuitively, was to train a student to mimic the teacher's outputs~\cite{bucila2006model}.
  This concept evolved into \textit{knowledge distillation}~\cite{hinton2015distilling}, which proposed using \textit{soft} probabilistic outputs, which are more informative than \textit{hard} class labels.
  These soft labels act as essential inter-class hints, conveying the hidden relationships that define an expert's logic.
  While foundational research has expanded these mechanisms, they generally assume a transparent environment where agents have full access to both original training data and the teacher model~\cite{romero2015fitnets, nguyen2021knowledge}.
\end{foldable}

\subsection{Constraints in Data Availability} \label{ssec:02-related-dfkd}
\begin{foldable} 
  When access to the primary data source is prohibited, researchers have explored data-efficient approaches.
  Few-shot KD methods such as~\cite{wang2020neural, nguyen2022black, vo2024improving} significantly reduce the required data volume but still rely on a small portion of the original source.
  Alternatively, other methods use proxy datasets to replace the source dataset~\cite{chen2021learning, tang2023distribution}.
  However, they often have a strong built-in bias that could be mismatched with the teacher's domain, as noted in~\cite{torralba2011unbiased}.
  Meanwhile, true data-free KD approaches often rely on dissecting the teacher's internal structure or leveraging its gradients as guidance to harvest knowledge~\cite{chen2019data, fang2021contrastive, do2022momentum, tran2024nayer}.
\end{foldable}

\subsection{Navigating Restricted Expert Interfaces} \label{ssec:02-related-bbdfkd}
\begin{foldable} 
  Still, data-free methods collapse when the expert operates as a black-box, returning only hard labels as a top\nobreakdash-1 prediction.
  The earliest method to address BBDFKD is ZSDB3~\cite{wang2021zero}, which assumes class robustness and image fidelity based on the distance to the decision boundary.
  This method uses random pixel-wise uniform noise for initialization and gradient estimation.
  However, high-dimensional boundaries can be over-complex for distance-based approaches, so obtaining diverse image samples scales exponentially worse with high-resolution images or numerous classes, limiting their success to small-scale, simple datasets.
\end{foldable}

\begin{foldable} 
  Subsequently, IDEAL~\cite{zhang2022ideal} and DFHL-RS~\cite{yuan2024data} utilize an extra generator network for more flexible data generation.
  Both assume that as the student improves during KD, its gradients could guide generator training.
  The methods employ in-class similarity and class-balance objectives for the generator, which is standard in data-free KD~\cite{chen2019data}.
  However, relying on the signals of an immature student is not reliable enough to train the generator.
  This setup has been observed to cause \textit{``overfitting of synthetic data''} to the student, resulting in possible mode collapse and insufficient diversity~\cite{yuan2024data}.
  Moreover, they are restricted to using hard labels instead of probabilities for KD, limiting the distillation signals.
\end{foldable}

\section{Framework} \label{sec:03-framework}
\begin{figure}[ht] 
  \centering
  \includegraphics[width=0.99\textwidth]{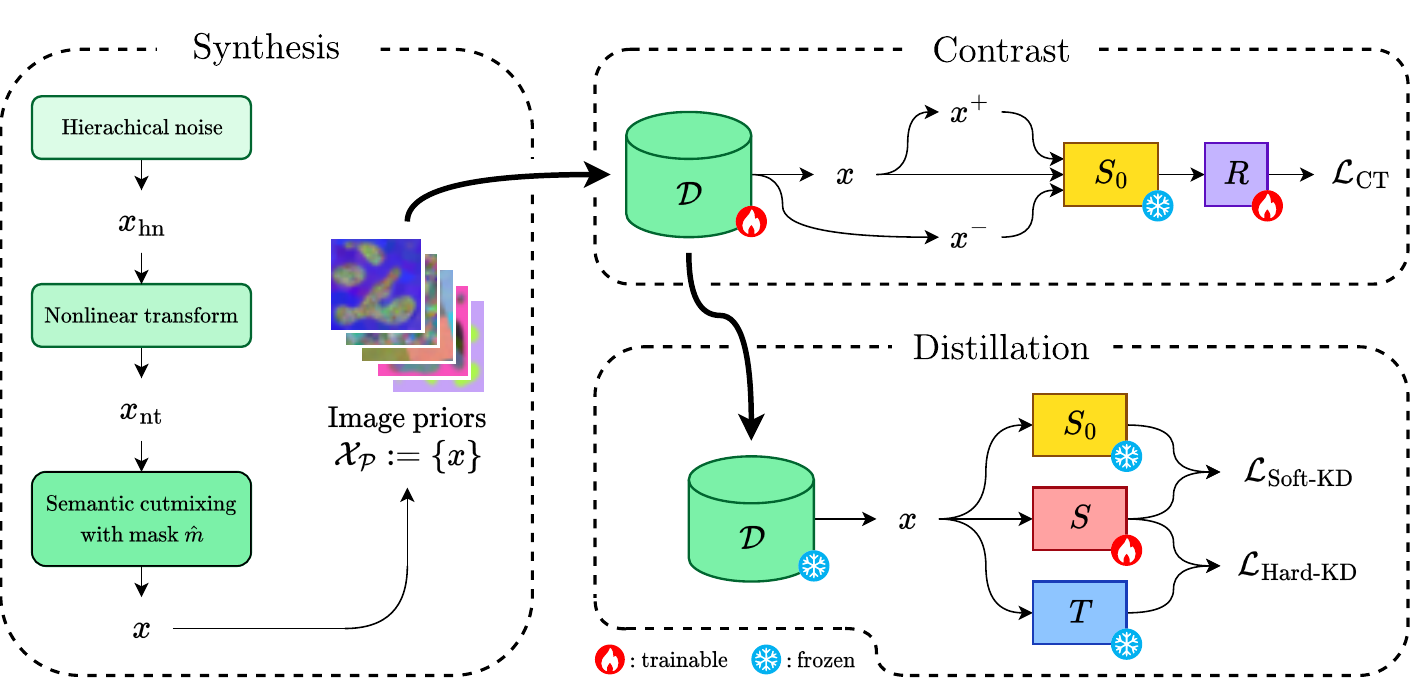}
  \caption{
    Illustration of our method \textbf{DIP-KD} of three phases.
    (\textbf{1})~\textbf{Synthesis}:{
      We craft image priors $\mathcal{X}_\mathcal{P}$ with three sub-components: hierarchical noise, nonlinear transformation, and semantic cutmixing.
      Synthetic images $x \sim \mathcal{X}_\mathcal{P}$ have diverse and distinct patterns.
    }
    (\textbf{2})~\textbf{Contrast}:{
      We construct the dataset $\mathcal{D} = \{x | x \sim \mathcal{X}_\mathcal{P}\}$ and train a primer student $S_0$ as a feature extractor.
      Then, we enforce an instance-discriminator $R$ to distinguish embeddings of images $x \sim \mathcal{D}$ to be similar to its positive view $x^{+}$, and dissimilar to its negative view $x^{-}\ne x$.
      Thus, $\mathcal{D}$ is optimized to be even more diverse.
    }
    (\textbf{3})~\textbf{Distillation}:{
      We distill the student $S$ by matching with both hard labels from teacher $T$ via $\mathcal{L}_\text{Hard-KD}$ and soft labels from primer student $S_0$ via $\mathcal{L}_\text{Soft-KD}$, providing extra knowledge.
    }
  }
  \label{fig:03-framework-teaser}
\end{figure}

\subsection{Problem setup} \label{ssec:03-framework-setup}
\begin{foldable} 
  \textbf{Problem statement.}
  Given solely a \textit{black-box} pre-trained teacher network $T$ that returns only top-1 hard labels (no internal features, logits, or gradients) and \textit{no real training data}, our goal is to train a student $S$ to approximate $T$.
\end{foldable}

\begin{foldable} 
  \textbf{Naive solution.}
  Lacking a direct solution, one might adapt standard KD~\cite{hinton2015distilling} as a naive baseline.
  Without real data, this necessitates using uniform noise images, \eg, $x \sim \mathcal{X}_\mathcal{U}$, where $\mathcal{X}_\mathcal{U} = \mathcal{U}[0,1]^{C\times H\times W}$.
  In this restricted interface, only the hard label $\hat{y}_{T}^\text{hard} = T(x) \in \{1, \ldots, K\}$ is available to guide the student's probabilistic outputs $\hat{y}_{S}^\text{soft} = S(x)$.
  The objective is defined as:
  \begin{equation} \label{eq:loss-naive-kd}
    \mathcal{L}_\text{NaiveKD} = \mathbb{E}_{x \sim \mathcal{X}_\mathcal{U}} \left[\mathcal{L}_\text{CE} \left(\hat{y}_{S}^\text{soft}, \hat{y}_{T}^\text{hard} \right)\right],
  \end{equation}
  where $\mathcal{L}_\text{CE}$ is the cross-entropy loss.
  While this can handle trivial cases, it apparently utilizes semantically non-diverse noise, and fails to leverage inter-class knowledge---the primary advantage of KD.
\end{foldable}

\begin{foldable} 
  \textbf{Proposal.}
  To improve BBDFKD, we propose \textbf{Diverse Image Priors Knowledge Distillation} (\textbf{DIP-KD}), structured into three collaborative phases:
  \begin{enumerate}
    \item {
        In the \textbf{Synthesis} phase, we design a pipeline to create \textit{image priors} $\mathcal{X}_\mathcal{P}$ that simulate the hierarchical structures, nonlinearity, and meaningful semantics in natural scenes and objects.
        This is achieved via three sub-components: hierarchical noise, nonlinear transformation, and semantic cutmixing.
      }
    \item {
        Then, in the \textbf{Contrast} phase, we sample a dataset $\mathcal{D} = \{x | x \sim \mathcal{X}_\mathcal{P}\}$.
        We further train a \textit{primer student} $S_0$ to optimize $\mathcal{D}$ via a contrastive learning objective to make every sample informationally \textit{distinguishable}.
      }
    \item {
        Finally, in the \textbf{Distillation} phase, we utilize the optimized dataset $\mathcal{D}$ to distill the student $S$.
        Crucially, we extract \textit{soft} knowledge from the primer student $S_0$ to restore the inter-class signals absent in previous BBDFKD methodologies.
      }
  \end{enumerate}
\end{foldable}

\subsection{Synthesis: Reconstructing the knowledge domain} \label{ssec:03-framework-synthesis}
\begin{figure}[ht] 
  \centering
  \includegraphics[width=0.99\textwidth]{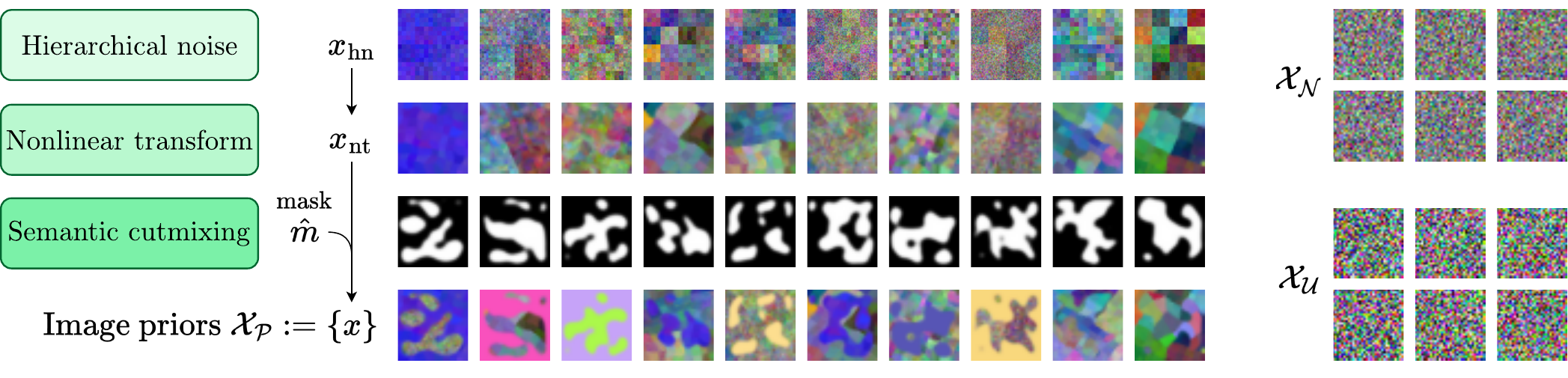}
  \caption{
    In the \textbf{Synthesis} phase, we generate image priors $\mathcal{X}_\mathcal{P}$ with 3 sub-components.
    (\textbf{1})~\textbf{Hierarchical noise}: We sample $x_\text{hn}$ from multi-scale noise images to capture hierarchical structures.
    (\textbf{2})~\textbf{Nonlinear transform}: We transform $x_\text{hn}$ to $x_\text{nt}$ with nonlinear filters to capture nonlinearity.
    and (\textbf{3})~\textbf{Semantic cutmixing}: We apply cutmixing on $x_\text{nt}$ with a semantic mask $\hat{m}$ to construct the final images $x$, simulating diverse semantics.
    Our image priors $\mathcal{X}_\mathcal{P}$ are visually much more diverse than noise images sampled from random Gaussian or random uniform distributions $\mathcal{X}_{\mathcal{N}}, \mathcal{X}_\mathcal{U}$.
  }
  \label{fig:03-framework-viz-synthesis}
\end{figure}

\begin{foldable} 
  Natural images are characterized by strong local pixel structures and complex, nonlinear semantics~\cite{lecun2002gradient}.
  To capture these properties, we generate image priors $\mathcal{X}_\mathcal{P}$ through a pipeline of three stages: \textit{hierarchical noise}, \textit{nonlinear transformation}, and \textit{semantic cutmixing}, visualized in Fig.~\ref{fig:03-framework-viz-synthesis}.
  This pipeline serves to capture the structural and semantic breadth of the expert's original knowledge domain through enhanced diversity.
\end{foldable}

\subsubsection{Hierarchical noise} \label{sssec:03-framework-synthesis-hiernoise}
\begin{foldable} 
  Hierarchy is a universal trait in natural scenes and objects.
  Based on this observation, we design a sampling technique that captures image patterns at both local and global scales to mimic the hierarchical structures of visual signals.
\end{foldable}

\begin{foldable} 
  In a standard pixel-wise noise image $x \sim X_\mathcal{U}$, patterns are primarily local and independent.
  Conversely, a monochromatic (single-color) image has dominant global correlation.
  To balance these patterns, we design a sampler that combines noise at multiple scales.
  For an image of dimension $C \times H \times W$, we sample noise images with increasing scales:
  \begin{equation} \label{eq:synthesis-hiernoise-multiscale}
    \left\{ x_{d} |x_{d} \sim \mathcal{U}[0,1]^{C\times2^{d}\times2^{d}} \right\}_{d=0}^{d_\text{max}},
  \end{equation}
  where $d_\text{max} = \text{ceil}(\log_{2} \max(H, W))$ is the first scale to sufficiently cover $H\times W$.
  The dimensions are derived deterministically, \eg, a $32 \times 32$ image is generated from scales $\{2^0, 2^1, \ldots, 2^5\}$.
  To combine local-to-global noise into a single image, we sample coefficients $\alpha_d$ from a normal distribution and apply softmax to get weights $\bar{\alpha}_{d}$.
  We then mix the noise images $\{x_d\}$ to obtain the \textit{hierarchical noise} image:
  \begin{equation} \label{eq:synthesis-hiernoise-combine}
    x_\text{hn} = \sum_{d=0}^{d_\text{max}} \bar{\alpha}_{d} \cdot f_\text{upscale}\left(x_{d};2^{d_\text{max}}\right),
  \end{equation}
  where $f_\text{upscale}$ upscales all noise images to the same dimension $2^{d_\text{max}}$ for proper mixing.
  This approach provides a multi-scale structural foundation that allows the student to learn complex visual hierarchies in the absence of real-world data.
\end{foldable}

\subsubsection{Nonlinear transformation} \label{sssec:03-framework-synthesis-nonlintrans}
\begin{foldable} 
  To enrich nonlinear patterns of synthetic images, we apply random rotation $f_\text{rot}$ of $[-45^{\circ}, 45^{\circ}]$, random elastic transform $f_\text{elas}$, and random cropping $f_\text{crop}$ to the specified size $H \times W$.
  These operations are known to be nonlinear and effective in data augmentation~\cite{simard2003best}, and in our case, boosting data diversity.
  Accordingly, we create a \textit{nonlinear-transformed} image as:
  \begin{equation} \label{eq:synthesis-nonlintrans}
    x_\text{nt} = f_\text{crop}^{H\times W} \circ f_\text{elas} \circ f_\text{rot} (x_\text{hn}).
  \end{equation}
\end{foldable}

\subsubsection{Semantic cutmixing} \label{sssec:03-framework-synthesis-semcutmix}
\begin{foldable} 
  We aim to augment synthetic images with the cohesive structures often found in natural objects.
  However, previous techniques like CutMix~\cite{yun2019cutmix} rely on simple rectangular occlusions that lack semantic meaning.
  To address this, we extend CutMix by improving the masking process to nonlinearly blend image pairs into \textit{semantically coherent} structures.
\end{foldable}

\begin{foldable} 
  We achieve these patterns through refinement of a semantic mask $\hat{m}$, initialized as $m_{0} \sim \mathcal{U}[0,1]^{1 \times H \times W}$.
  To balance spatial continuity with regional partitioning, we respectively use a blurring filter $f_\text{blur}$ and our \textit{diverging filter} $f_\text{divg}$.
  Specifically, $f_\text{divg}$ is a piecewise quadratic operator that is continuously differentiable at its inflection point (0.5), driving pixel values toward binary states (0 or 1), producing sharp, well-defined region boundaries.
  \begin{equation} \label{eq:synthesis-diverging-filter}
    f_\text{divg}(m) =
    \begin{cases}
      2 m^2           & , m \ge 0.5 \\
      1 - 2 (m - 1)^2 & , m < 0.5
    \end{cases}.
  \end{equation}
\end{foldable}

\begin{foldable} 
  We obtain the final $\hat{m}$ by iteratively applying $m_{i+1} = f_\text{divg} \circ f_\text{blur}(m_i)$ until the mean absolute pixel-wise update reaches a steady-state threshold $\epsilon < 10^{-2}$, empirically around 10 iterations.
  This mask is used for pairwise cutmixing $\{x_\text{nt}\}$ against themselves or random monochromatic images $\mathcal{X}_\text{mc}$ to form our final \textit{image priors} $\mathcal{X}_\mathcal{P} = \{x\}$, where:
  \begin{equation} \label{eq:synthesis-semcutmix}
    x= \text{CutMix}(x_i, x_j; \hat{m}), \text{\quad s.t. \quad} x_i,x_j \sim \{x_\text{nt}\} \cup \mathcal{X}_\text{mc} \text{ and } x_i \ne x_j.
  \end{equation}
  This approach ensures the synthetic domain of $\mathcal{X}_\mathcal{P}$ captures diverse semantics required for effective black-box knowledge acquisition.
\end{foldable}

\subsection{Contrast: Collaborative diversity optimization} \label{ssec:03-framework-contrast}
\begin{foldable} 
  Standard contrastive learning methods require internal feature extraction, which is prohibited by the black-box nature of our teacher.
  We propose the novel use of a \textit{primer student} $S_{0}$ as a white-box mediator to bridge this informational gap.
  We first train $S_{0}$ on an image priors dataset $\mathcal{D} = \{x | x \sim \mathcal{X}_\mathcal{P}\}$ following the Hard-KD objective:
  \begin{equation} \label{eq:loss-hard-kd}
    \mathcal{L}_\text{Hard-KD} = \mathbb{E}_{x \sim \mathcal{D}} \left[\mathcal{L}_\text{CE} \left(\hat{y}_{S_0}^\text{soft}, \hat{y}_{T}^\text{hard} \right)\right],
  \end{equation}
  where $\hat{y}_{S_0}^\text{soft} = S_0(x)$ is its probability prediction, and $\hat{y}_{T}^\text{hard} = T(x)$ is the teacher's hard label.
  Equipped with this white-box proxy, we can access its internal representations for explicit diversity optimization.
\end{foldable}

\begin{foldable} 
  Following~\cite{fang2021contrastive}, diversity correlates with how \textit{``distinguishable''} samples are, under an \textit{explicitly optimizable} objective.
  To maximize diversity, we use the primer student's backbone $S_{0}^\mathcal{H}$ to extract image embeddings and append an instance-discriminator $R$ (a shallow MLP) to recognize each embedding as a unique class.
  We argue that even if $S_0$ is yet to match the teacher's expertise, its backbone $S_{0}^\mathcal{H}$ is capable of capturing the underlying semantics required for contrastive learning.
  This is supported by prior works in contrastive learning~\cite{chen2020simple,he2020momentum,zbontar2021barlow}, which do not depend on a high-accuracy classifier, but often train from scratch.
  For any two images $(x_i, x_j)$, we compute their cosine similarity as:
  \begin{equation}
    \text{Sim}\left(x_i, x_j\right)=
    \frac{ \left\langle R \circ S_{0}^\mathcal{H}(x_i), R \circ S_{0}^\mathcal{H}(x_j)\right\rangle }
    { \left\Vert R \circ S_{0}^\mathcal{H}(x_i) \right\Vert \cdot \left\Vert R \circ S_{0}^\mathcal{H}(x_j) \right\Vert }.
  \end{equation}
\end{foldable}

\begin{foldable} 
  For diversity optimization, from an image $x \sim \mathcal{D}$, we generate positive views $x^{+}$ via augmentation and sample other images $x^{-} \sim \mathcal{D}, x^{-} \neq x$ as negative views.
  The contrastive objective maximizes the similarity between $(x, x^{+})$ and minimizes that between $(x, x^{-})$:
  \begin{equation} \label{eq:loss-contrastive}
    \mathcal{L}_\text{CT} = \mathbb{E}_{x \sim \mathcal{D}} \left[
      -\log \frac{ \exp\left( \text{Sim}(x, x^{+}) \right) }{ \sum_{x^{-} \ne x} \exp\left( \text{Sim}(x, x^{-}) \right) }
    \right].
  \end{equation}
  By backpropagating $\nabla\mathcal{L}_\text{CT}$ to update both the instance-discriminator $R$ and the samples $x \in \mathcal{D}$, we explicitly reinforce the semantic diversity of $\mathcal{D}$.
\end{foldable}

\subsection{Distillation: Restoring informative signals} \label{ssec:03-framework-distillation}
\begin{foldable}
  Once the synthetic dataset $\mathcal{D}$ has been optimized for diversity, we use it for the final training of the student $S$.
  Our objective is to restore the high-bandwidth signals typical of standard KD~\cite{hinton2015distilling}, \ie, incorporating both \textit{hard} and \textit{soft} knowledge, which were \textit{limited or entirely absent} in recent works~\cite{wang2021zero, zhang2022ideal, yuan2024data}.
  Having privileged access to the primer network $S_0$, we form an optimal distillation objective by jointly matching the student's logits $S^\mathcal{Z}(x)$ with soft labels $S_{0}^\mathcal{Z}(x)$ and hard labels $\hat{y}_{T}^\text{hard}$:
  \begin{equation} \label{eq:loss-kd-total}
    \mathcal{L}_{S}
    = \mathcal{L}_\text{Hard-KD} + \mathcal{L}_\text{Soft-KD} \\
    = \mathbb{E}_{x \sim \mathcal{D}}\Big[
      \mathcal{L}_\text{CE}\left( \hat{y}_{S}^\text{soft}, \hat{y}_{T}^\text{hard} \right) +
      \mathcal{L}_\text{L1}\left( S^\mathcal{Z}(x), S_{0}^\mathcal{Z}(x) \right)
    \Big],
  \end{equation}
  where $\mathcal{L}_\text{L1}$ is the L1 divergence loss, applied directly to logits to preserve raw inter-class relationships.
  By reconciling the teacher's authoritative expertise with the primer student's relational hints, the student $S$ overcomes the signal scarcity of the black-box interface.
  This collaborative training allows for a deeper understanding of the teacher's decision-making logic, resulting in superior performance across restricted AI ecosystems.
\end{foldable}

\section{Experiments} \label{sec:04-experiments}
\begin{foldable} 
  We evaluate DIP-KD on extensive BBDFKD benchmarks, supplemented by ablation studies to validate its mechanics and practical effectiveness.
\end{foldable}

\subsection{Experimental setup} \label{ssec:04-experiments-setup}
\begin{foldable} 
  We evaluate our method across 12 benchmarks, including 8 general-purpose datasets (USPS, MNIST, SVHN, FMNIST, CIFAR10, CIFAR100, Tiny-ImageNet, Imagenette)~\cite{lecun2002gradient,hull1994database,netzer2011reading,xiao2017fashion,krizhevsky2009learning,le2015tiny,howard2020imagenette} and 4 subsets from the medical dataset MedMNIST~\cite{yang2021medmnist}.
  We investigate three teacher-student architectures: LeNet5~\cite{lecun2002gradient}, AlexNet~\cite{krizhevsky2012imagenet}, and ResNet~\cite{he2016deep}.
  These datasets and architectures are the de facto evaluation in BBDFKD methods~\cite{wang2021zero, zhang2022ideal, yuan2024data}.
\end{foldable}

\subsection{Baselines} \label{ssec:04-experiments-baselines}
\begin{foldable} 
  We compare our method DIP-KD with the baselines:
  \begin{itemize}
    \item \textbf{NaiveKD}: The student is trained on uniform noise $x \sim \mathcal{X}_\mathcal{U}$, where $\mathcal{X}_\mathcal{U}=\mathcal{U}[0,1]^{C \times H \times W}$ and the teacher's hard labels $\hat{y}_{T}^\text{hard} = T(x) \in \{1, \ldots, K\}$.
    \item \textbf{BBDFKD methods}: We reproduce three state-of-the-art methods: ZSDB3~\cite{wang2021zero}, IDEAL~\cite{zhang2022ideal}, and DFHL-RS~\cite{yuan2024data}.
  \end{itemize}
  For fair comparison, we use the same teacher-student network architectures and synthetic image budget for all methods.
  We report the \textit{mean} $\pm$ \textit{standard error} (SE) of distillation accuracy, which is the student accuracy on the hold-out test set, across five runs.
  The accuracies of baseline BBDFKD methods are reproduced with their official implementations\footnote{ZSDB3 at \url{https://github.com/zwang84/zsdb3kd}}\,\footnote{IDEAL at \url{https://github.com/SonyResearch/IDEAL}}\,\footnote{DFHL-RS at \url{https://github.com/LetheSec/DFHL-RS-Attack}}.
  We ensured that all baseline implementations were tuned ethically and fairly following the recommended settings.
\end{foldable}

\subsection{Standard experiments} \label{ssec:04-experiments-standard}
\begin{foldable} 
  We categorize the benchmarks as \textit{simple}, \textit{complex}, and \textit{domain-specific} datasets.
  Four simple datasets---MNIST, USPS, SVHN, FMNIST---all have 10 classes of simple objects, at a $32\times32$ or smaller resolution.
  The complex datasets have more complicated objects or more classes: CIFAR10 and CIFAR100 (10 and 100 classes, $32\times32$), Tiny-ImageNet (200 classes, $64\times64$), and Imagenette (10 classes, but high-resolution).
  Finally, we evaluate the domain-specific performance on 4 subsets of MedMNIST, including BreastMNIST, OrganAMNIST, DermaMNIST, and BloodMNIST, which include $32\times32$ images of varying medical modalities.
  We report the distillation accuracy of all methods in Table~\ref{tab:standard-all}, where our method \textbf{DIP-KD} consistently outperforms all baselines across 11/12 datasets, with significant margins on complex and domain-specific ones.
\end{foldable}

\begin{table}[ht]
  \centering
  \begin{threeparttable}
    \caption{Distillation accuracy (\%) on simple, complex, and domain-specific datasets.}
    \label{tab:standard-all}
    \small
    \begin{tabular}{l c c c c c c c}
      \toprule
      \boldtight{Dataset} & \boldtight{Setup} & \boldtight{Teacher} & \boldtight{NaiveKD} & \boldtight{ZSDB3} & \boldtight{IDEAL} & \boldtight{DFHL-RS} & \boldtight{DIP-KD} \\
      \midrule

      \multicolumn{8}{c}{\textbf{Simple datasets}} \\
      \midrule
      MNIST    & L, 20K  & 99.28 & 96.54$_{\pm 0.60}$           & 96.66$_{\pm 0.39}$           & 96.96$_{\pm 0.34}$ & 98.53$_{\pm 0.10}$          & \textbf{98.59}$_{\pm 0.08}$ \\
      USPS     & L, 50K  & 95.47 & 42.05$_{\pm 0.08}$           & 65.50$_{\pm 0.16}$           & 92.66$_{\pm 0.57}$ & 93.97$_{\pm 0.60}$          & \textbf{94.20}$_{\pm 0.21}$ \\
      SVHN     & A, 50K  & 96.16 & 83.35$_{\pm 0.57}$           & 85.38$_{\pm 1.69}$           & 87.86$_{\pm 0.93}$ & 95.41$_{\pm 0.13}$          & \textbf{95.94}$_{\pm 0.05}$ \\
      FMNIST   & L, 50K  & 90.90 & 55.11$_{\pm 1.23}$           & 71.60$_{\pm 2.35}$           & 82.84$_{\pm 1.22}$ & 83.89$_{\pm 1.35}$          & \textbf{83.98}$_{\pm 0.65}$ \\
      \midrule

      \multicolumn{8}{c}{\textbf{Complex datasets}} \\
      \midrule
      CIFAR10  & R, 50K  & 95.21 & 13.99$_{\pm 0.23}$           & 56.74$_{\pm 1.98}$           & 67.58$_{\pm 1.29}$ & 76.50$_{\pm 0.84}$          & \textbf{83.41}$_{\pm 0.46}$ \\
      CIFAR100 & R, 100K & 78.36 & \phantom{0}2.68$_{\pm 0.13}$ & \phantom{0}9.19$_{\pm 0.39}$ & 33.64$_{\pm 0.91}$ & 51.97$_{\pm 0.11}$          & \textbf{52.85}$_{\pm 0.30}$ \\
      TinyImgN & R, 500K & 66.08 & \phantom{0}0.75$_{\pm 0.07}$ & \phantom{0}3.29$_{\pm 0.16}$ & 23.76$_{\pm 1.03}$ & \textbf{30.52}$_{\pm 1.43}$ & 28.03$_{\pm 0.22}$ \\
      Imagenette & R, 50K  & 91.29 & 33.21$_{\pm 0.39}$         & 33.43$_{\pm 0.85}$           & 36.90$_{\pm 0.38}$ & 40.53$_{\pm 2.36}$          & \textbf{61.84}$_{\pm 0.77}$ \\
      \midrule

      \multicolumn{8}{c}{\textbf{Domain-specific datasets}} \\
      \midrule
      BreastM  & L, 50K  & 79.49 & 61.78$_{\pm 1.60}$           & 63.37$_{\pm 1.48}$           & 67.54$_{\pm 1.27}$ & 73.81$_{\pm 1.80}$          & \textbf{77.95}$_{\pm 0.77}$ \\
      OrganAM  & L, 50K  & 87.60 & 24.09$_{\pm 0.44}$           & 51.81$_{\pm 1.21}$           & 69.09$_{\pm 1.92}$ & 75.12$_{\pm 2.89}$          & \textbf{82.32}$_{\pm 1.08}$ \\
      DermaM   & R, 50K  & 79.60 & 56.82$_{\pm 1.50}$           & 60.15$_{\pm 2.93}$           & 62.76$_{\pm 2.76}$ & 64.71$_{\pm 3.02}$          & \textbf{71.29}$_{\pm 2.67}$ \\
      BloodM   & R, 50K  & 98.19 & \phantom{0}9.09$_{\pm 0.13}$ & 39.62$_{\pm 1.67}$           & 55.28$_{\pm 2.18}$ & 61.67$_{\pm 3.11}$          & \textbf{73.98}$_{\pm 1.77}$ \\
      \bottomrule
    \end{tabular}
    \begin{tablenotes}[flushleft] \footnotesize
    \item {Setup denotes architecture \{\textbf{\underline{L}}eNet5, \textbf{\underline{A}}lexNet, \textbf{\underline{R}}esNet18\} and synthetic dataset size; [\textbf{bold}] best results.}
    \end{tablenotes}
  \end{threeparttable}
\end{table}

\begin{foldable}
  While performance saturates across all baselines on simple datasets, the accuracy gap widens as domain complexity scales.
  NaiveKD is first to collapse to near random-guessing levels in complex datasets, implying that unstructured noise cannot elicit the deep semantic knowledge embedded in black-box experts.
  Notably, on high-resolution images like Imagenette, our DIP-KD achieves a +21.31\% absolute gain over the strongest baseline, DFHL-RS.
  On domain-specific medical tasks, DIP-KD outperforms baselines by at least +4\% to +12\%.
  We also notice that abundant classes represent an interesting boundary condition, where DIP-KD remains highly competitive on CIFAR100 and approaches state-of-the-art on Tiny-ImageNet.
  These results suggest that data diversity is a critical factor for effective knowledge acquisition, particularly as task complexity and domain specificity increase.
\end{foldable}

\begin{foldable}
  In terms of robustness, we observe that DIP-KD has consistently low standard errors on most datasets, indicating strong stability across runs.
  We attribute this robustness to our integrated focus on diversity and signal restoration, which provide the student with softer training signals than noise-heavy or hard-label baselines.
\end{foldable}

\subsection{Ablation studies} \label{ssec:04-experiments-ablation}
\begin{foldable}
  In this section, we provide deeper insights into our framework through component dissection and practical application.
  We first investigate the contribution of individual components, the effect of synthetic data budget, and the semantics of our image priors.
  Then, we further extend our evaluation to real-world scenarios, including cross-architecture settings and aggressive model compression.
\end{foldable}

\subsubsection{Contribution of components} \label{sssec:04-experiments-ablation-components}

\begin{table}[ht]
  \centering
  \begin{threeparttable}
    \caption{Distillation accuracy (\%) by inclusion of components.}
    \label{tab:abla-components}
    \begin{tabular}{l l l}
      \toprule
      \makecell{\textbf{Method}} & \makecell{\textbf{MNIST}} & \makecell{\textbf{CIFAR10}} \\
      \midrule
      NaiveKD (noise $x \sim \mathcal{X}_\mathcal{U}$, Hard-KD) & \phantom{$+$ }\meanstd{96.02}{.60} & \phantom{$+$ }\meanstd{13.99}{.23} \\
      $+$ \textbf{Synthesis:}                                   & $+$ \phantom{0}2.27                & $+$ 64.18 \\
      \phantom{12} \bltrg Hierarchical noise                    & \phantom{$+$ 0}1.79\bltrg          & \phantom{$+$} 59.26\bltrg  \\
      \phantom{12} \bltrg Nonlinear transform                   & \phantom{$+$ 0}0.18\bltrg          & \phantom{$+$ 0}2.38\bltrg  \\
      \phantom{12} \bltrg Semantic cutmixing                    & \phantom{$+$ 0}0.30\bltrg          & \phantom{$+$ 0}2.54\bltrg  \\
      $+$ \textbf{Contrast:} Update $\mathcal{D}$ via $\mathcal{L}_\text{CT}$ & $+$ \phantom{0}0.16  & $+$ \phantom{0}3.14 \\
      $+$ \textbf{Distillation:} Add Soft-KD to $\mathcal{L}_S$ & $+$ \phantom{0}0.14                & $+$ \phantom{0}2.10 \\
      \midrule
      Complete \textbf{DIP-KD}                                  & \phantom{$+$ }98.59$_{\pm .08}$    & \phantom{$+$ }83.41$_{\pm .46}$ \\
      \bottomrule
    \end{tabular}
    \begin{tablenotes}[flushleft] \footnotesize
    \item {[$\blacktriangle$] denotes individual contribution of each Synthesis sub-component.}
    \end{tablenotes}
  \end{threeparttable}
\end{table}

\begin{foldable}
  We evaluate each component on MNIST and CIFAR10 to represent simple and complex benchmarks.
  Starting from NaiveKD, we sequentially add components and report student accuracy in Table~\ref{tab:abla-components}.
  Synthesis provides the largest gain ($+$2.27\% on MNIST, $+$64.18\% on CIFAR10), primarily driven by hierarchical noise.
  This is equivalent to the primer student $S_0$ trained on $\mathcal{X}_\mathcal{P}$, achieving 98\% and 78\% on MNIST and CIFAR10 (which already outperforms the baselines from Sec. \ref{ssec:04-experiments-standard}).
  This confirms that $S_0$ is sufficiently competent to perform contrastive optimization and soft distillation.
  For instance, on CIFAR10, we achieve $+$3.14\% via Contrast and $+$2.10\% via Distillation.
  These results confirm that the Synthesis pipeline plays a vital role in generating diverse data, while the novel use of the primer student for Contrast and Distillation phases significantly boosts KD performance beyond prior works.
\end{foldable}

\subsubsection{Embeddings analysis --- Synthesis} \label{sssec:04-experiments-ablation-embeddings-generation}
\begin{figure}[ht] 
  \centering
  \includegraphics[trim={120 0 0 0},clip,width=0.90\columnwidth]{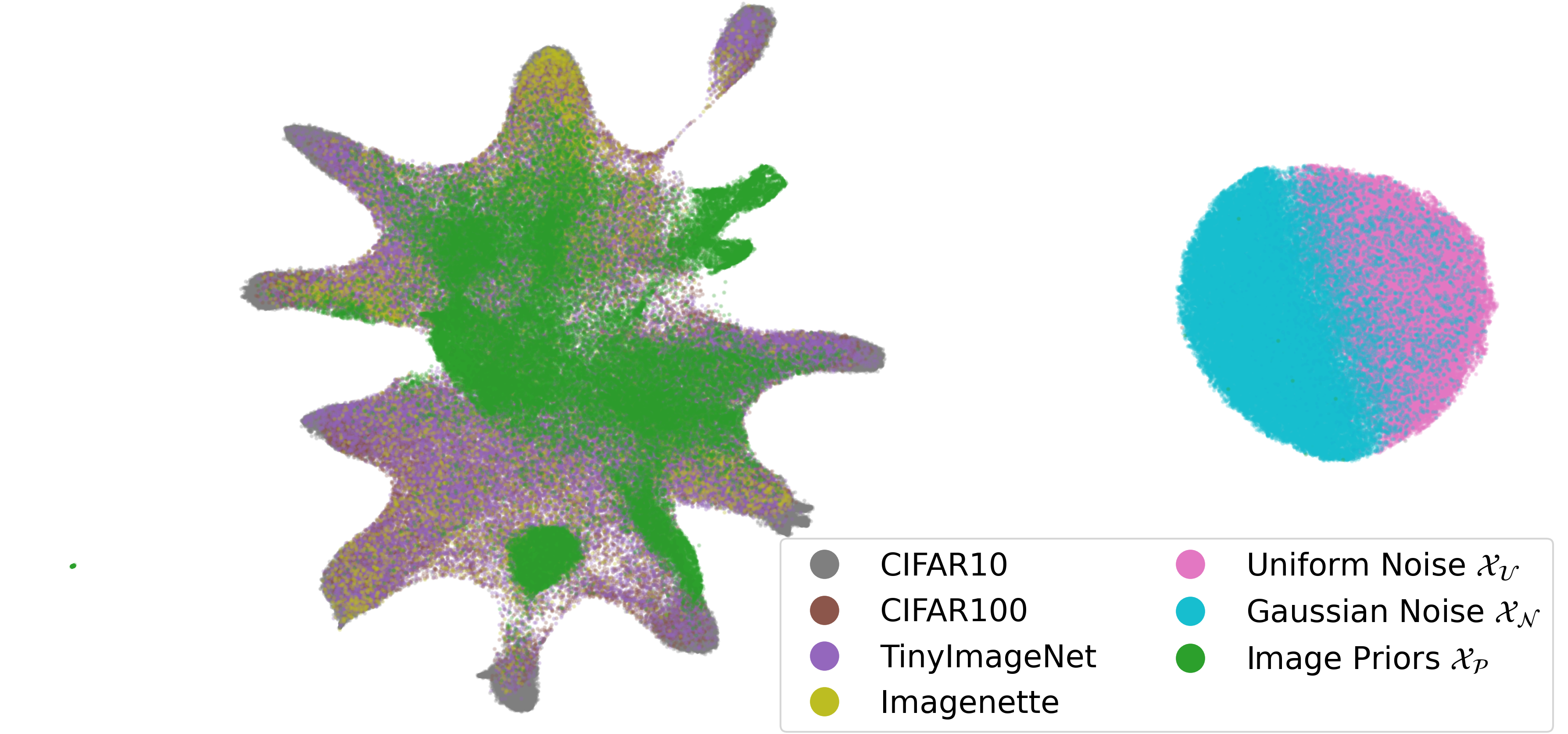}
  \caption{Feature embeddings of complex datasets, noise ($\mathcal{X}_\mathcal{U}$, $\mathcal{X}_\mathcal{N}$), and image priors $\mathcal{X}_\mathcal{P}$.}
  \label{fig:embeddings-universal}
\end{figure}

\begin{foldable}
  To assess the diversity and semantics of our image priors, we sample synthetic images and extract embeddings from a ResNet18 teacher backbone pre-trained on CIFAR10.\footnote{This is the 95.21\% ResNet18 teacher on CIFAR10, being reused as a held-out ``oracle''.}
  These 512-dimensional features are reduced to two dimensions using UMAP~\cite{mcinnes2018umap} with default configuration.
  We explore the remaining complex datasets CIFAR100, Tiny-ImageNet, and Imagenette, as well as uniform noise $\mathcal{X}_\mathcal{U}$, Gaussian noise $\mathcal{X}_\mathcal{N}$, and our image priors $\mathcal{X}_\mathcal{P}$.
  We declare that the use of real data and the teacher's backbone here is solely for evaluation purposes.
  We visualize these embeddings in Fig.~\ref{fig:embeddings-universal}.
\end{foldable}

\begin{table}[ht]
  \centering
  \begin{threeparttable}
    \caption{Distribution metrics (\%) on complex datasets.}
    \label{tab:abla-embeddings-coverage-objects}
    \small
    \begin{tabular}{c c c c c c}
      \toprule
      \textbf{Metric} & \textbf{Source} & \textbf{CIFAR10} & \textbf{CIFAR100} & \textbf{Tiny-ImageNet} & \textbf{Imagenette} \\
      \midrule
      & $\mathcal{X}_\mathcal{U}$ & \phantom{0}20.00 &           100.65 &           108.65 &           101.20 \\
      \textbf{Density}   & $\mathcal{X}_\mathcal{N}$ & \phantom{0}20.00 &           107.98 & \phantom{0}90.87 &           103.64 \\
      & $\mathcal{X}_\mathcal{P}$ & \phantom{0}49.14 & \phantom{0}96.20 & \phantom{0}96.00 & \phantom{0}92.14 \\
      \midrule
      & $\mathcal{X}_\mathcal{U}$ & \phantom{000}.00 & \phantom{000}.10 & \phantom{000}.06 & \phantom{000}.05 \\
      \textbf{Coverage}  & $\mathcal{X}_\mathcal{N}$ & \phantom{000}.00 & \phantom{000}.10 & \phantom{000}.06 & \phantom{000}.05 \\
      & $\mathcal{X}_\mathcal{P}$ & \phantom{00}7.33 & \phantom{0}60.79 & \phantom{0}63.88 & \phantom{0}78.48 \\
      \midrule
      & $\mathcal{X}_\mathcal{U}$ &           100.00 &           100.10 &           100.00 &           100.00 \\
      \textbf{Precision} & $\mathcal{X}_\mathcal{N}$ &           100.00 &           100.10 &           100.00 &           100.00 \\
      & $\mathcal{X}_\mathcal{P}$ & \phantom{0}80.83 & \phantom{0}97.76 & \phantom{0}99.29 & \phantom{0}99.10 \\
      \midrule
      & $\mathcal{X}_\mathcal{U}$ & \phantom{000}.00 & \phantom{000}.09 & \phantom{000}.06 & \phantom{000}.04 \\
      \textbf{Recall}    & $\mathcal{X}_\mathcal{N}$ & \phantom{000}.00 & \phantom{000}.10 & \phantom{000}.06 & \phantom{000}.05 \\
      & $\mathcal{X}_\mathcal{P}$ & \phantom{0}66.92 & \phantom{0}96.75 & \phantom{0}96.86 & \phantom{0}95.07 \\
      \bottomrule
    \end{tabular}
  \end{threeparttable}
\end{table}

\begin{foldable}
  As shown, $\mathcal{X}_\mathcal{U}$ and $\mathcal{X}_\mathcal{N}$ form a compact, separate cluster far from real images, indicating that their semantics are limited and dissimilar to natural objects.
  In contrast, $\mathcal{X}_\mathcal{P}$ overlaps with real image regions, suggesting that they share similar semantics.
  For quantitative analysis, we evaluate four distribution metrics (precision, recall, density, coverage) from~\cite{naeem2020reliable}.
  The statistics in Table~\ref{tab:abla-embeddings-coverage-objects} show that our image priors $\mathcal{X}_\mathcal{P}$ achieve significantly higher density, coverage, and recall at a small trade-off in precision, compared to noise sources $\mathcal{X}_\mathcal{U}$ and $\mathcal{X}_\mathcal{N}$.
  These results further confirm that $\mathcal{X}_\mathcal{P}$ has superior fidelity and diversity, and explain why $\mathcal{X}_\mathcal{P}$ excels in BBDFKD.
\end{foldable}

\subsubsection{Embeddings analysis --- Contrast} \label{sssec:04-experiments-ablation-embeddings-contrast}
\begin{figure}[ht] 
  \centering
  \includegraphics[trim={120 0 0 0},clip,width=0.90\columnwidth]{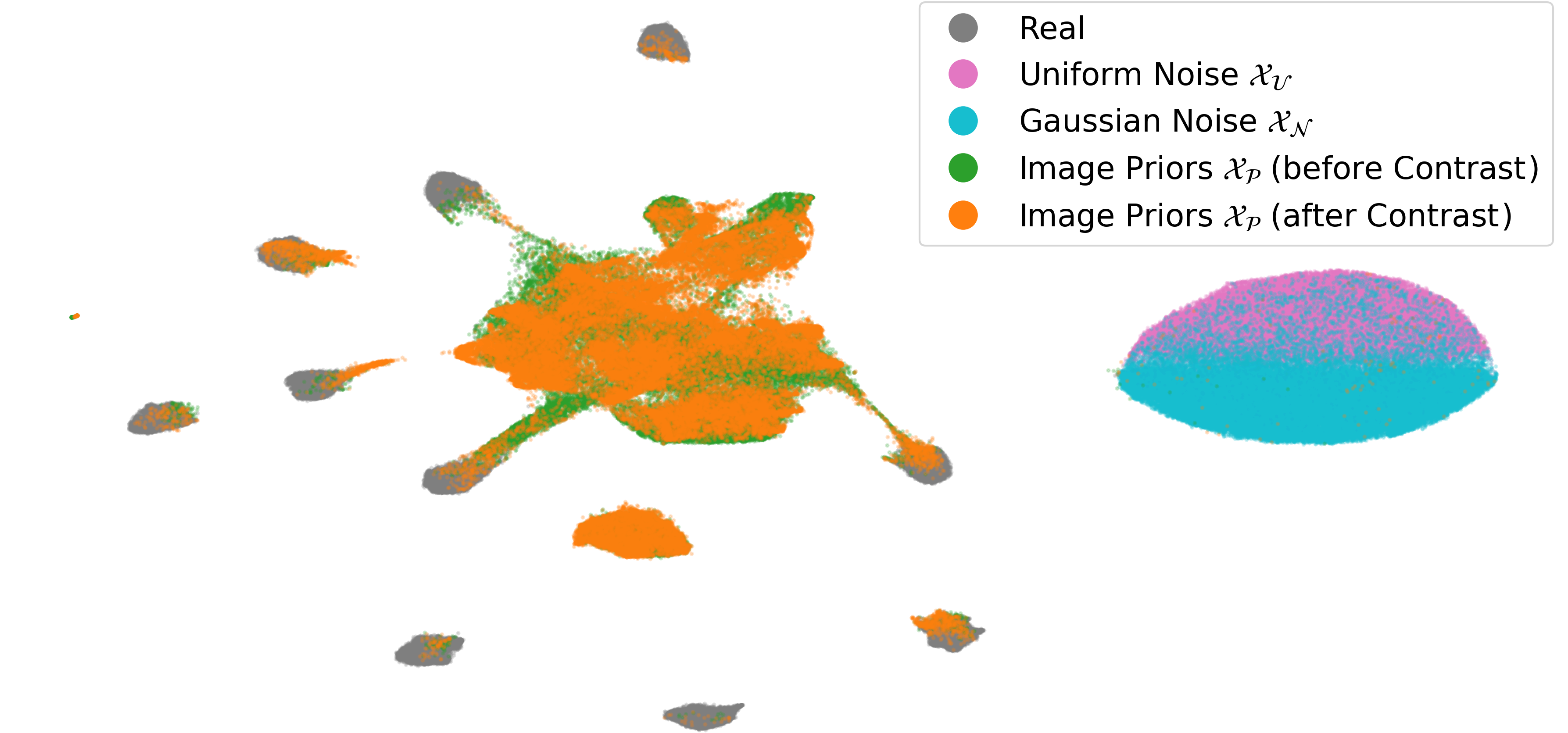}
  \caption{Feature embeddings of image priors $\mathcal{X}_\mathcal{P}$ before and after the Contrast phase.}
  \label{fig:embeddings-cifar10}
\end{figure}

\begin{foldable}
  To analyze the Contrast phase's impact on diversity, we evaluate 50K image prior embeddings in Fig.~\ref{fig:embeddings-cifar10}.
  The backbone embeds real images into ten clusters (equivalent CIFAR10 classes; gray).
  While pre-Contrast priors $\mathcal{X}_\mathcal{P}$ (green) already scatter around real images, the Contrast phase (orange) expands coverage into neighboring regions.
  This increased semantic diversity lets the student capture broader teacher knowledge.
  The distribution metrics in Table~\ref{tab:abla-embeddings-coverage-cifar10} confirm this with a significant increase in coverage and recall, with a trivial trade-off in density.
  Overall, these results show that through fulfilling the contrastive objective, the Contrast phase improves data diversity semantically and empirically.
\end{foldable}

\begin{table}[ht]
  \centering
  \begin{threeparttable}
    \caption{Distribution metrics (\%) during the Contrast phase.}
    \label{tab:abla-embeddings-coverage-cifar10}
    \begin{tabular}{l c c c c}
      \toprule
      & \textbf{Density} & \textbf{Coverage} & \textbf{Precision} & \textbf{Recall} \\
      \midrule
      Before Contrast & 70.46            & \phantom{0}6.13   & 91.69              & 82.18 \\
      After Contrast  & 66.93            &           14.66   & 91.68              & 94.00 \\
      \bottomrule
    \end{tabular}
  \end{threeparttable}
\end{table}

\subsubsection{Increasing synthetic dataset size} \label{sssec:04-experiments-ablation-nsamples}
\begin{figure}[H] 
  \centering
  \includegraphics[width=0.80\columnwidth]{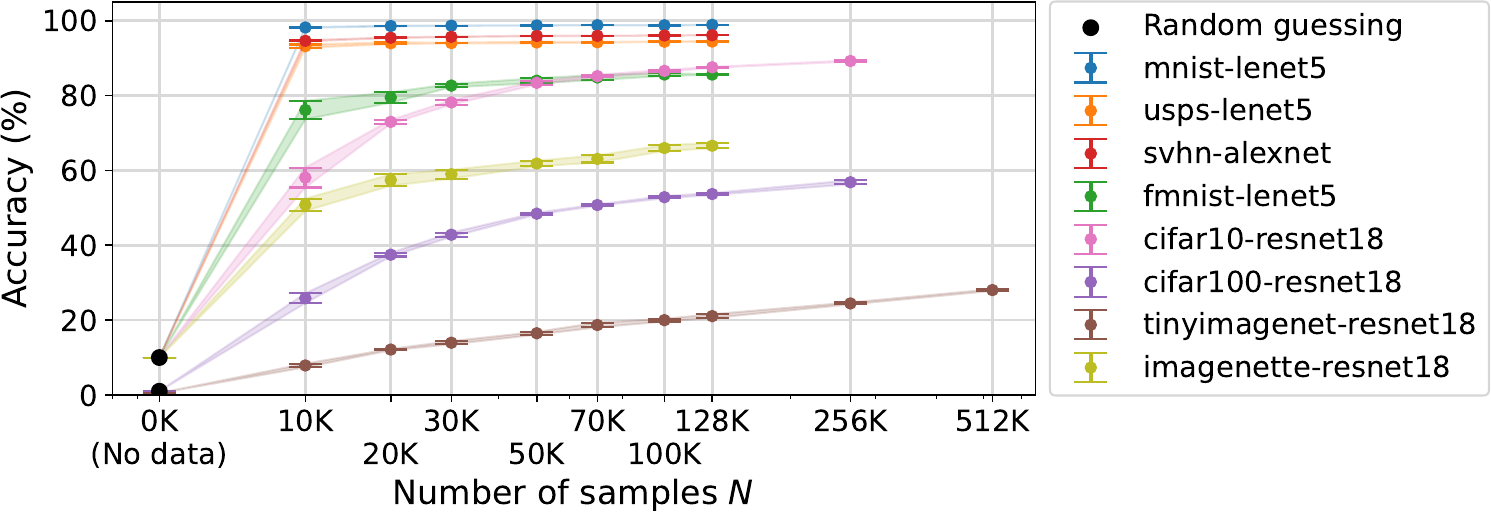}
  \caption{Distillation accuracy (\%) by number of image priors $N$.}
  \label{fig:abla-nsamples}
\end{figure}

\begin{foldable}
  The relationship between synthetic data budget $N$ and KD performance mirrors the emergence of collective intelligence.
  While it is expected that KD performance improves with large $N$ (Fig.~\ref{fig:abla-nsamples}), we are interested in the optimal cut-off point, which indicates the \textit{minimum quorum} required for the student to replicate the teacher's expertise.
  This threshold scales directly with the structural complexity of the environment.
  For digits datasets (MNIST, USPS, SVHN), minimal data ($N$ = 10K) is required to reach good results, implying informational redundancy.
  In large-scale settings (CIFAR100, Tiny-ImageNet, Imagenette), the student continues to improve around $N$ = 100K, suggesting that the information density has not yet saturated and the collective wisdom of the teacher's priors remains beneficial.
\end{foldable}

\subsubsection{Cross-architecture KD} \label{sssec:04-experiments-ablation-crossarch}
\begin{table}[ht]
  \centering
  \caption{Distillation accuracy (\%) of cross-architecture KD.}
  \label{tab:abla-crossarch}
  \begin{threeparttable}
    \begin{tabular}{l l l l}
      \toprule
      & \textbf{FMNIST} & \textbf{SVHN} & \textbf{CIFAR10} \\
      \midrule
      Teacher  & LeNet5 & AlexNet & ResNet18 \\
      & 90.90 & 96.16 & 95.21 \\
      \midrule
      LeNet5   & \meanstdbf{83.98}{0.65} & \meanstd{64.44}{1.83} & \meanstd{32.01}{3.18} \\
      AlexNet  & \meanstd{78.98}{0.92} & \meanstdbf{95.94}{0.05} & \meanstd{61.30}{0.69} \\
      ResNet18 & \meanstd{81.69}{1.58} & \meanstd{95.91}{0.04} & \meanstdbf{83.41}{0.46} \\
      \bottomrule
    \end{tabular}
    \begin{tablenotes}[flushleft] \footnotesize
    \item {[\textbf{bold}] best results.}
    \end{tablenotes}
  \end{threeparttable}
\end{table}

\begin{foldable}
  An under-investigated setting in BBDFKD methods is when the teacher and student have different architectures.
  We evaluate KD with heterogeneous student architectures (LeNet5, AlexNet, ResNet18) on FMNIST, SVHN, and CIFAR10.
  Table~\ref{tab:abla-crossarch} shows that students perform best when matching the teacher's architecture (diagonal), with performance degrading as the capacity gap widens as expected.
  Thus, DIP-KD is practical in real-world scenarios with unknown teachers, provided the student has sufficient capacity.
\end{foldable}

\subsubsection{Students of compressed capacity} \label{sssec:04-experiments-ablation-compress}
\begin{figure}[ht] 
  \centering
  \includegraphics[width=0.80\columnwidth]{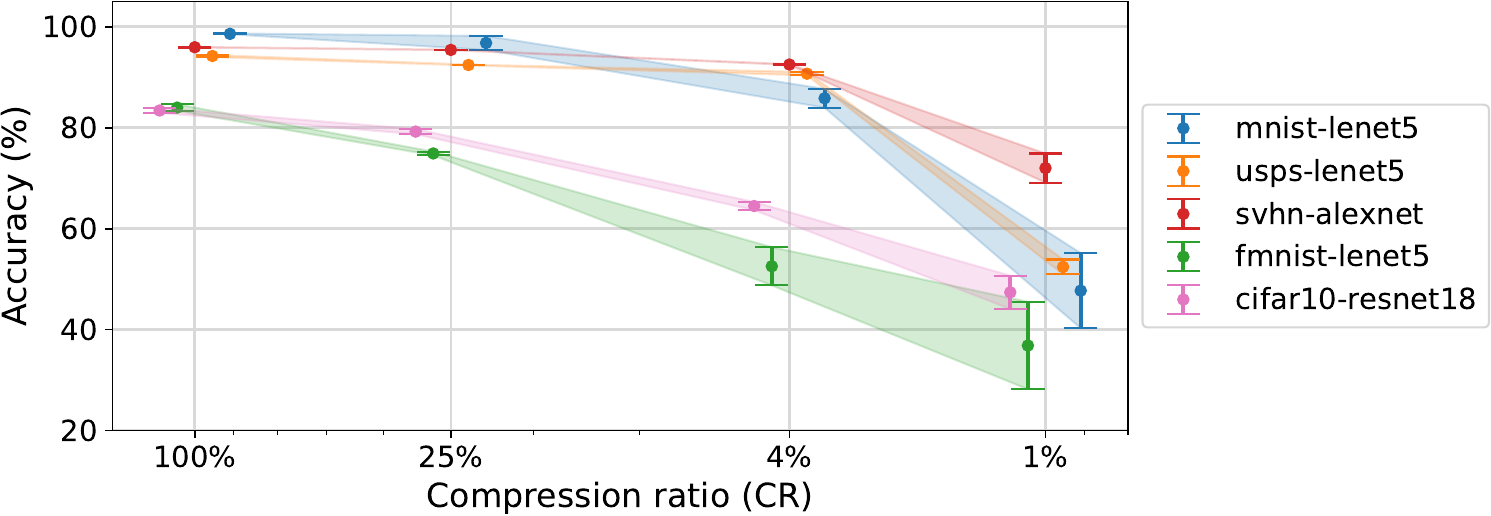}
  \caption{Distillation accuracy (\%) by compression ratio (CR).}
  \label{fig:abla-compression}
\end{figure}

\begin{foldable}
  Beyond same-size configurations, we evaluate DIP-KD under aggressive compression, a scenario unaddressed in prior methods~\cite{wang2021zero,zhang2022ideal,yuan2024data}.
  Specifically, we distill student networks with 25\%, 4\%, and 1\% compression ratio (CR) by reducing channel counts by 2$\times$, 5$\times$, 10$\times$, respectively.
  Results indicate strong empirical robustness: performance remains stable at 25\% CR, with only tolerable decline at 4\% CR.
  Significant degradation occurs only at 1\% CR, where the student's extreme sparsity prevents it from capturing the complex inter-class signals.
  These results demonstrate that DIP-KD is ideal for \textit{resource-constrained} mobile or edge environments.
\end{foldable}

\section{Conclusion} \label{sec:05-conclusion}
This work addresses information flow bottlenecks in decentralized AI systems, where privacy and proprietary barriers create a black-box data-free environment.
We introduce \textbf{DIP-KD}, a framework framing KD as the emergence of intelligence from a collective of synthetic image priors.
By coordinating a three-phase pipeline: Synthesis, Contrast, and Distillation, we enable students to emulate expert logic through highly restricted interfaces.
We achieve state-of-the-art performance across 12 benchmarks, with notable improvements in complex and domain-specific settings.
Beyond establishing a new empirical baseline, this study identifies data diversity as a key factor for effective knowledge acquisition in restricted environments.
We hope our work can provide a robust foundation for future research into resilient and decentralized collective intelligence.

\bibliography{references}

\end{document}